\pdfoutput=1

\documentclass[11pt]{article}

\usepackage[preprint]{acl}

\usepackage{times}
\usepackage{latexsym}

\usepackage[T1]{fontenc}

\usepackage[utf8]{inputenc}

\usepackage{microtype}

\usepackage{inconsolata}

\usepackage{graphicx}

\usepackage{booktabs}

\newcommand\rqChange{How much do raters change prior annotations?}
\newcommand\rqDetect{Do raters detect artificial error spans?}

\newcommand\rqHumanQuality{Does re-annotating human annotations improve their quality?}
\newcommand\rqAutoQuality{Do automatic prior annotations improve human annotation quality?}

\title{MQM Re-Annotation: A Technique for Collaborative Evaluation of Machine Translation}

\author{Parker Riley~~~{\bf Daniel Deutsch}~~~{\bf Mara Finkelstein} \\ {\bf Colten DiIanni}~~~{\bf Juraj Juraska}~~~{\bf Markus Freitag} \\ Google}

\begin{document}
\maketitle
\begin{abstract}
Human evaluation of machine translation is in an arms race with translation model quality: as our models get better, our evaluation methods need to be improved to ensure that quality gains are not lost in evaluation noise. To this end, we experiment with a two-stage version of the current state-of-the-art translation evaluation paradigm (MQM), which we call MQM re-annotation. In this setup, an MQM annotator reviews and edits a set of pre-existing MQM annotations, that may have come from themselves, another human annotator, or an automatic MQM annotation system. We demonstrate that rater behavior in re-annotation aligns with our goals, and that re-annotation results in higher-quality annotations, mostly due to finding errors that were missed during the first pass. 

\end{abstract}

\section{Introduction}

A critical component of machine translation (MT) model development is evaluating model quality. Despite extensive work on automatic evaluation \citep{guerreiro-etal-2024-xcomet,juraska-etal-2024-metricx,freitag-etal-2024-llms,kocmi-federmann-2023-gemba,fernandes-etal-2023-devil,rei-etal-2020-comet}, human evaluation remains the gold standard for reliably determining quality. However, there are still issues affecting the quality of human evaluation, such as differences in rater behavior and task difficulty. These can negatively affect the reliability of human evaluation, and as the quality of MT systems improves, there is a risk that this evaluation noise can outweigh actual quality differences between models, leading to incorrect modeling decisions. As a result, human evaluation must also be improved to facilitate comparing ever-higher-quality systems \citep{freitag-etal-2021-experts, kocmi-etal-2024-findings, kocmi-etal-2023-findings}.

The current state-of-the-art human evaluation framework for MT is Multidimensional Quality Metrics (MQM), originally proposed by \citet{lommel-etal-mqm}. \citet{freitag-etal-2021-experts} adapted MQM for MT, arguing that the use of fine-grained annotation by expert raters improved the quality of human evaluation. In MQM, expert raters identify error spans within a translation, assigning a severity and hierarchical category to each. Follow-up work has examined other facets of evaluation that can be used to enhance quality, including comparative judgment \citep{song-etal-2025-enhancing}, careful rater assignment \citep{riley-etal-2024-finding}, and rater calibration \citep{kocmi-etal-2022-findings}. 

\begin{figure}
    \centering
    \includegraphics[width=0.95\linewidth]{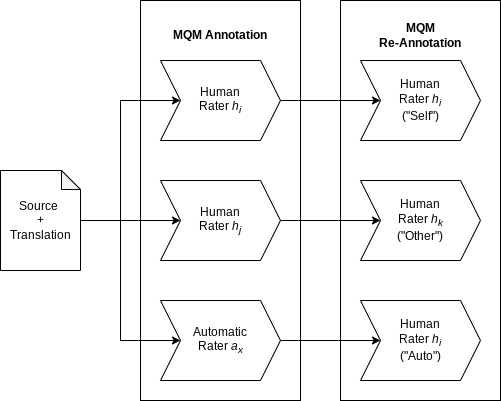}
    \caption{Illustration of MQM re-annotation. A source document and its translation are annotated in the MQM framework by either a human or an automatic system, and then re-annotated by a human. If the initial annotator was a human, the re-annotator can be either the same person or a different one.}
    \label{fig:re_anno_intro}
\end{figure}

In this work we explore both human-machine collaboration and human-human collaboration, with the goal of increasing quality of MQM annotation for machine translation evaluation. We do this using \textit{re-annotation}, where a completed MQM annotation of a document is given to an expert human rater who can delete or modify any previously-marked error spans in addition to adding new ones. The prior rater can either be the same human rater, a different human rater, or an automatic system. Figure~\ref{fig:re_anno_intro} illustrates these three cases.

\section{Research Questions and Findings}

In this section, we outline the research questions we seek to answer in this work and summarize the findings. Evidence supporting these findings is presented in Section~\ref{sec:results}, with relevant context and background covered in the intervening sections.

\subsection{\rqChange}
Our first goal is to understand what happens when raters perform MQM re-annotation: how often do they change or delete prior errors, and on average how many new ones do they add? Further, to what extent does this depend on the source of those prior errors (the same rater, a different human rater, or an automatic system)?

We find that rater behavior is highly sensitive to the source of the prior errors, and trends align with our expectations: raters perform the fewest overall modifications to their own annotations, more to other humans' annotations, and even more modifications to automatic annotations. We also find evidence suggesting that a primary benefit of re-annotation is that it allows raters the opportunity to find errors that were missed initially.
\subsection{\rqDetect} 

One potential risk with re-annotation is that raters may overly trust the prior errors and focus solely on adding new ones. To examine this, we injected artificial error spans and measured how often raters modified or deleted them.

We found that, on average, raters removed most of the artificial errors, but that a minority of raters accepted these errors as-is with high frequency.

\subsection{\rqHumanQuality} 
The prior two research questions establish that re-annotation is \textit{reasonable}, in that re-annotators behave in ways that are consistent with our expectations, but we also wish to understand whether it is \textit{beneficial}, in that it improves rating quality. Because defining a ground truth is difficult in human evaluation research, we measure whether re-annotation can allow two disjoint groups of human raters to achieve higher cross-group agreement.

We find that re-annotating human MQM ratings (either by the same rater or a different human rater) increases their quality.

\subsection{\rqAutoQuality} 
As a cheaper alternative to human-human re-annotation, we examine whether re-annotating \textit{automatic} ratings is beneficial over standard MQM annotation from scratch. Based on the previous research question, we can use our re-annotated data as the proxy ground truth to measure improvement.

We find that providing raters with prior annotations from high-quality LLM-based automatic systems improves rating quality over from-scratch annotation, at no additional human annotation cost.

\section{MQM and Re-Annotation}

In the Multidimensional Quality Metrics (MQM) framework for human evaluation of machine translation (MT), expert annotators identify error spans within translations and assign a hierarchical category (e.g. Fluency/Grammar or Accuracy/Mistranslation) and severity (Major or Minor). Instead of raters providing scalar scores, scores are calculated by applying a weighting scheme based primarily on severity to each error and summing over the segment; see \citet{freitag-etal-2021-experts} for details. All source and target segments for a single system's translation of a single document are presented to the rater on the same screen.

MQM re-annotation functions the same as standard MQM annotation, except that some error spans are already present when raters first load the interface. Raters have the ability to modify any component of a prior error (span start/end, category, severity), delete it entirely, and add new errors. 

\begin{figure}
    \centering
    \includegraphics[width=0.5\textwidth]{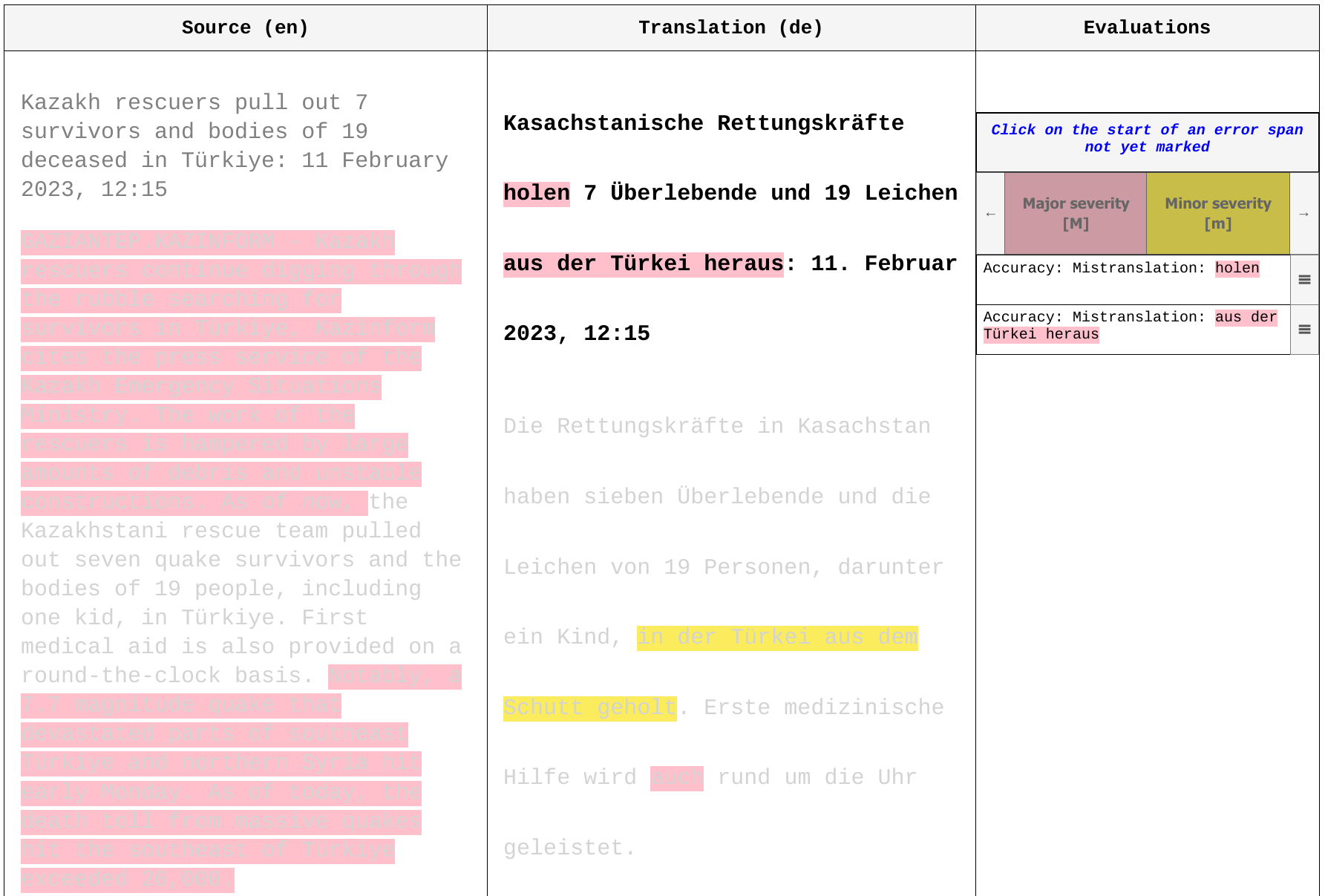}
    \caption{Anthea interface showcasing prior error annotations.}
    \label{fig:interface}
\end{figure}

We use the open-source Anthea\footnote{https://github.com/google-research/google-research/tree/master/anthea} framework to collect MQM ratings, including re-annotations. The rating interface is shown in Figure~\ref{fig:interface}.

\section{Experimental Setup}
\subsection{Data}
We use translation data from the WMT 2023 Metrics Shared Task \citep{freitag-etal-2023-results} in two language pairs: English-German and Chinese-English. For budget reasons, we only use data from the News domain. Information about the dataset size is shown in Table~\ref{tab:data_stats}.

The data contains 3 initial human MQM annotations per segment, and two automatic annotations from different systems (Section~\ref{sec:automatic_systems}). Each of these annotations was then re-annotated by one of the same 3 human raters who initially rated that segment. Further details are in Section~\ref{sec:re_anno_settings}.

The human annotators are professional translators who regularly perform MQM annotation. They were paid fair market wages for their work.

\begin{table}[htb]
\centering
\footnotesize
\begin{tabular}{lrr}
\toprule
& Chinese-English & English-German \\
\midrule
\# Source Segments & 247 & 100 \\
\# Source Documents & 25 & 29 \\
\# Systems & 16 & 13 \\
\# Raters & 8 & 10 \\
\bottomrule
\end{tabular}
\caption{Counts of segments, documents, and systems in each language pair. One system was a human reference.}\label{tab:data_stats}
\end{table}

\subsection{Re-Annotation Settings}\label{sec:re_anno_settings}

We use $h_i,~i \in [1,N]$, to refer to MQM annotations from our $N$ human raters for a particular language pair, with no re-annotation. We use $a_x,~x \in \{1,2\}$ to refer to MQM annotations produced by an automatic system. For re-annotation, we adopt notation from conditional probability: for example, we use $h_i|h_j$ to refer to the MQM annotations from human rater $i$ when re-annotating those from human rater $j$.

We categorize (re-)annotations into the following types:
\begin{itemize}
\item \textbf{Single} refers to initial annotations by a single human rater\footnote{We do not consider automatic annotations in isolation in this work.}: $h_i,~i \in [1,N]$
\item \textbf{Self} refers to re-annotation done by the same human rater who did the initial annotation: $h_i|h_i$
\item \textbf{Other} refers to re-annotation done by a human rater who is different from the initial rater: $h_j|h_i,~i \neq j$
\item \textbf{Auto} refers to re-annotation done by a human rater when the initial annotations came from an automatic system: $h_i|a_x$
\end{itemize}

\begin{table}[htb]
\centering
\footnotesize
\begin{tabular}{lrr}
\toprule
& Chinese-English & English-German \\
\midrule
Single & 11856 & 3900 \\
Self & 3952 & 1300 \\
Other & 7904 & 2600 \\
Auto & 7410 & 2400 \\
\bottomrule
\end{tabular}
\caption{Number of segment annotations for each \\(re-)annotation setting and language pair.}\label{tab:setting_stats}
\end{table}

Due to budget constraints, we were unable to collect data from all possible combinations of raters for every source document. Each document was rated by 3 human raters and 2 automatic systems, and these initial annotations were distributed among the same 3 human raters as follows: one rater was randomly selected to re-annotate their own ratings (the Self setting), and the other two re-annotated each other's ratings (the Other setting). Independently, 2 of the 3 raters were randomly selected to each re-annotate a different automatic system. All random assignments were made separately for each document, and each document's assignment was used for all system outputs of all segments of that document. However, automatic annotations were not available for the human reference system in the data (``refA''), so the Auto setting is somewhat smaller than the Other setting. When re-annotating, raters were not told the origin of the initial annotations. Table~\ref{tab:setting_stats} details how many segment annotations our data contains for each setting. 

\subsection{Automatic Systems}\label{sec:automatic_systems}
We experiment with two automatic MQM annotation systems: GEMBA-MQM \citep{kocmi-federmann-2023-gemba}, which is based on prompting GPT-4 \citep{openai2024gpt4technicalreport}, and a variant of AutoMQM \citep{fernandes-etal-2023-devil} based on fine-tuning Gemini 1.0 Pro \citep{geminiteam2024geminifamilyhighlycapable}. The AutoMQM model was fine-tuned on MQM data from the WMT Metrics Shared Task, years 2020 through 2022 \citep{mathur-etal-2020-results,freitag-etal-2021-results,freitag-etal-2022-results}. Both models are reference-free, taking as input only the source segment and corresponding translation hypothesis.

\subsection{Artificial Spans}
To test for over-reliance on prior annotations, we selected one document in each language pair to augment with artificial error spans. For each system translation of this document, and for each segment, we randomly selected a one- or two-token span that did not overlap with any errors in the three human initial annotations and assigned a Major severity and a random category. The Major severity was chosen to increase our confidence that the artificial errors were indeed incorrect. These artificial spans were combined with the actual spans identified by the rater who marked the most total errors for the chosen document, summed over all system translations. This was done to maximize the number of legitimate error spans in order to make the artificial spans less obvious. The chosen document was only used to analyze rater behavior on the artificial spans, and was not included in any other analyses in this work.

\subsection{Agreement Metrics}\label{sec:metrics}
\subsubsection{Score Agreement}
To measure inter-annotator agreement of MQM scores, we use Pairwise Ranking Agreement (PRA) \citep{song-etal-2025-enhancing,deutsch-etal-2023-ties}. PRA essentially measures how often two metrics (in our case, two sets of human annotations) agree on the relative ranking of two translations, treated as a 3-way classification task: A is either better than, worse than, or equal to B. Details can be found in \citet{song-etal-2025-enhancing}. We use the Group-by-Item setup of \citet{deutsch-etal-2023-ties}, where PRA is calculated over all system translations of a given source segment, and the average of this value over all source segments is reported.

\subsubsection{Span Agreement}
To measure inter-annotator agreement of error spans and severity markings, we use character-level F1 scores (abbreviated ``character F1''), introduced by \citet{blain-etal-2023-findings}. Character F1 involves comparing whether two annotations agree on whether a character is or isn't included in an error span, with half-points awarded when both annotations include the character in a span but disagree on the severity.

\section{Results}\label{sec:results}

\subsection \rqChange

\begin{table*}[htb]
\centering
\footnotesize
\begin{tabular}{lrrrrrr}
\toprule
& \multicolumn{3}{c}{Chinese-English} & \multicolumn{3}{c}{English-German}\\
& Self & Other & Auto & Self & Other & Auto\\
\midrule
Deleted \% & 10.9 & 25.3 & 28.7 & 10.2 & 20.9 & 24.9\\
Changed \% & 7.8 & 12.8 & 23.3 & 6.4 & 11.7 & 27.6\\
Kept \% & 81.3 & 61.9 & 48.0 & 83.4 & 67.3 & 47.5\\
Added \% & 30.1 & 59.2 & 173.8 & 41.0 & 54.6 & 202.1\\
\bottomrule
\end{tabular}
\caption{Number of deleted/changed/kept/added errors, expressed as percentages of the total number of prior errors, by language pair and re-annotation setting. Deleted \%, Changed \%, and Kept \% add to 100\% within a column. Added \% can be higher than 100\% if the number of newly added errors exceeds the number of prior errors. Percentages are macro-averages of individual re-annotator performances.}\label{tab:change_rates}
\end{table*}

Table~\ref{tab:change_rates} reports the relative frequencies of deleting, changing, keeping, and adding errors during re-annotation. In both language pairs, non-keep rates increase when changing setting from Self to Other to Auto. These results validate our hypothesis that any given rater should generally agree with themselves more than a different human, and with a different human more than with an automatic rater. The Added \% for the Auto setting is strikingly higher than in the other two settings, leading us to compare the average total number of errors identified over the entire dataset by our human and automatic raters. We found that human annotators identified 2.7 times as many errors during initial annotation as automatic systems did, in both language pairs.

Additionally, we note that re-annotation increases the average number of annotated error spans in all settings. This is true even in the Self setting, where one might assume that raters would change very little from their initial annotation. This provides evidence that a primary benefit of re-annotation is \textbf{finding missed errors} (as opposed to removing/changing incorrectly-annotated errors).

\subsection \rqDetect

\begin{table}[htb]
\centering
\footnotesize
\begin{tabular}{lrr}
\toprule
& Chinese-English & English-German\\
\midrule
Deleted \% & 82.3 & 70.8\\
Changed \% & 4.9 & 4.4\\
Kept \% & 12.8 & 24.8\\
\bottomrule
\end{tabular}
\caption{Percentage of artificial error annotations that were deleted/changed/kept by re-annotators. Percentages are macro-averages of individual re-annotator performances.}\label{tab:artificial_change_rates}
\end{table}
Table~\ref{tab:artificial_change_rates} shows the rates at which re-annotators deleted, changed, or kept artificial error spans. While re-annotators usually did not keep artificial error spans unchanged, they did so more than we expected. We expected the percentage of kept errors to be nearly 0 because we selected artificial error spans that did not overlap with any prior human annotations. When examining per-rater performance, we found that a few outlier raters were keeping prior errors at concerningly high rates (78.8\% for one English-German rater). The median percentage of kept artificial errors is 8.8\% for Chinese-English and 13.5\% for English-German.

\subsection \rqHumanQuality

\begin{table*}[htb]
\centering
\footnotesize
\begin{tabular}{lrrrrrrrr}
\toprule
& \multicolumn{4}{c}{Chinese-English} & \multicolumn{4}{c}{English-German}\\
& $h_j$ & $h_j|h_k$ & $h_k$ & $h_k|h_j$ & $h_j$ & $h_j|h_k$ & $h_k$ & $h_k|h_j$\\
& (Single) & (Other) & (Single) & (Other) & (Single) & (Other) & (Single) & (Other) \\
\midrule
$h_i$ (Single) & 0.306 & 0.344 & 0.322 & 0.354 & 0.324 & 0.362 & 0.348 & 0.352\\
$h_i|h_i$ (Self) & 0.321 & 0.359 & 0.340 & 0.372 & 0.346 & 0.394 & 0.364 & 0.384\\

\bottomrule
\end{tabular}
\caption{Character F1 scores between different annotation settings.}\label{tab:human_cf1}
\end{table*}

\begin{table*}[htb]
\centering
\footnotesize
\begin{tabular}{lrrrrrrrr}
\toprule
& \multicolumn{4}{c}{Chinese-English} & \multicolumn{4}{c}{English-German}\\
& $h_j$ & $h_j|h_k$ & $h_k$ & $h_k|h_j$ & $h_j$ & $h_j|h_k$ & $h_k$ & $h_k|h_j$\\
& (Single) & (Other) & (Single) & (Other) & (Single) & (Other) & (Single) & (Other) \\
\midrule
$h_i$ (Single) & 0.526 & 0.549 & 0.536 & 0.551 & 0.572 & 0.613 & 0.576 & 0.612\\
$h_i|h_i$ (Self) & 0.534 & 0.566 & 0.549 & 0.573 & 0.590 & 0.659 & 0.613 & 0.646\\

\bottomrule
\end{tabular}
\caption{Average segment-level pairwise agreement scores between different annotation settings.}\label{tab:human_acc23}
\end{table*}

Tables~\ref{tab:human_cf1} and~\ref{tab:human_acc23} report the agreement between different (re-)annotation settings using character F1 and average segment-level PRA, respectively (see Section~\ref{sec:metrics} for details on agreement metrics). Under both metrics, the agreement between two annotations in the Single setting (e.g. $h_i$ vs. $h_j$) is increased by replacing either side with a re-annotation (e.g. $h_i|h_i$ vs. $h_j$), and increased further by replacing both sides (e.g. $h_i|h_i$ vs. $h_j|h_k$). Because all agreement scores are calculated between annotations stemming from two disjoint sets of human annotators (avoiding bias), we conclude that the increased agreement is an indication that re-annotation improves the quality of human annotations.

\subsection \rqAutoQuality

\begin{table*}[htb]
\centering
\footnotesize
\begin{tabular}{lrrrrrr}
\toprule
& \multicolumn{3}{c}{Chinese-English} & \multicolumn{3}{c}{English-German}\\
& $h_j$ & $h_j|a_y$ & $h_j|h_k$ & $h_j$ & $h_j|a_y$ & $h_j|h_k$\\
& (Single) & (Auto) & (Other) & (Single) & (Auto) & (Other) \\
\midrule
$h_i$ (Single) & 0.315 & 0.330 & 0.332 & 0.353 & 0.383 & 0.368\\
$h_i|a_x$ (Auto) & 0.352 & 0.380 & 0.375 & 0.411 & 0.422 & 0.428\\
$h_i|h_i$ (Self) & 0.337 & 0.337 & 0.348 & 0.384 & 0.401 & 0.412\\
\bottomrule
\end{tabular}
\caption{Character F1 scores between different annotation settings.}\label{tab:auto_cf1}
\end{table*}

\begin{table*}[htb]
\centering
\footnotesize
\begin{tabular}{lrrrrrr}
\toprule
& \multicolumn{3}{c}{Chinese-English} & \multicolumn{3}{c}{English-German}\\
& $h_j$ & $h_j|a_y$ & $h_j|h_k$ & $h_j$ & $h_j|a_y$ & $h_j|h_k$\\
& (Single) & (Auto) & (Other) & (Single) & (Auto) & (Other) \\
\midrule
$h_i$ (Single) & 0.511 & 0.510 & 0.530 & 0.574 & 0.598 & 0.579\\
$h_i|a_x$ (Auto) & 0.530 & 0.546 & 0.552 & 0.619 & 0.631 & 0.630\\
$h_i$|$h_i$ (Self) & 0.528 & 0.530 & 0.553 & 0.598 & 0.632 & 0.641\\
\bottomrule
\end{tabular}
\caption{Average segment-level pairwise agreement scores between different (re-)annotations.}\label{tab:auto_acc23}
\end{table*}

Having established that re-annotating human ratings improves their quality, we can use our re-annotations as proxies for the unknown ``ground truth'' MQM annotations and investigate whether providing raters with automatic prior annotations improves their rating quality. Tables~\ref{tab:auto_cf1} and~\ref{tab:auto_acc23} show average segment-level agreement between (re-)annotation settings, with a goal of determining whether the Auto setting ($h_i|a_x$, $h_j|a_y$) improves over the Single setting ($h_i$, $h_j$) when using either the Self ($h_i|h_i$) or Other ($h_j|h_k$) setting as the ground truth. Note that, in order to ensure that all comparisons within a language pair use the exact same data, these tables use 2/3 of the available data, discarding the third where neither of the initial ratings from the automatic systems were distributed to the rater who re-annotated their own ratings ($h_i$). In both language pairs, under both metrics, and for both $h_i|a_x$ and $h_j|a_y$, we see that providing humans with automatic prior annotations improves their agreement with human-only re-annotation, indicating an increase in quality; the only exception is in Table~\ref{tab:human_cf1} for Chinese-English where $h_j|a_y$ ties $h_j$ in terms of span-aware agreement with $h_i|h_i$.

Critically, these results illustrate that automatic prior annotations can increase human rating quality without any additional human annotation cost.\footnote{We found that re-annotation of automatic systems took 12\% less time than initial annotation for Chinese-English and 2\% less time for English-German, on average.}

\section{Related Work}

While our work focuses on improving evaluation quality, a notable recent work focusing instead on evaluation cost is Error Span Annotation (ESA) \citep{kocmi-etal-2024-error}. ESA is highly similar to MQM, but the primary cost-saving difference is the removal of error category information. This reduces annotation time, and \citet{kocmi-etal-2024-error} claim that it also permits the use of non-expert raters.

The most closely related work to ours is that of \citet{zouhar-etal-2025-ai}, who essentially do re-annotation of automatic prior annotations in the ESA framework. The largest difference between their work and ours is that we also investigate re-annotation of \textit{human} prior annotations, which we show can be used to create higher-quality reference annotations. It also provides a novel, direct way to examine inter-annotator agreement by directly comparing re-annotators' behavior on their own ratings to their behavior on others'. Additionally, we expect that our use of expert annotators reduces the available headroom for improving rating quality, making the observed increases from re-annotation (both of human and automatic prior annotations) more noteworthy.

Other works have attempted to increase the quality of machine translation evaluation, including MQM. \citet{song-etal-2025-enhancing} leveraged side-by-side MQM annotation to improve rating consistency. \citet{riley-etal-2024-finding} demonstrated that careful rater assignment can mitigate noise from differences in rater behavior. \citet{knowles-2021-stability} showed that score normalization, a technique intended to address differences in rater behavior, can be harmful if applied in the wrong setting. \citet{saldias-fuentes-etal-2022-toward} proposed a technique for selecting test data to send for human evaluation that improved reliability in a budget-constrained setting.

Re-annotation is conceptually similar to translation post-editing, where the translation itself is updated in a two-stage process as opposed to the \textit{annotation} of the translation as in this work. In this way, our work is loosely reminiscent of \citet{liu-etal-2024-beyond-human}, who investigated human-only and human-machine collaboration for translation post-editing.

\section{Conclusion}
In this work, we presented an analysis of MQM re-annotation, exploring settings where the initial annotator was the re-annotator themselves, a different human, or an automatic system. We validated that the re-annotation task elicits reasonable rater behavior, and provided evidence that it leads to higher-quality annotations. In the case of re-annotation of automatic ratings, this quality improvement comes at no additional human annotation cost. As for re-annotation of human ratings, the improved quality makes it a candidate for creating test sets for meta-evaluation of automatic metrics, especially span-level metrics where the quality of the ground-truth spans is critical for reliable hill-climbing.
\section*{Limitations}
All annotation and re-annotation in this work was exclusively performed by professional translators who regularly do MQM annotation, meaning that our findings may not apply to less-experienced annotators. Additionally, we only experimented with two automatic MQM systems that both have reasonable quality; it is possible that re-annotating a lower-quality system would not outperform from-scratch human-only annotation. 
Finally, our data is exclusively from the News domain, leaving the possibility that re-annotators might perform differently on different kinds of text.

\bibliography{anthology,custom}

\end{document}